%% file: acl_latex.tex
\documentclass[11pt]{article}

\usepackage[preprint]{acl}

\usepackage{times}
\usepackage{latexsym}
\usepackage{amsmath}
\usepackage{amssymb}
\usepackage{mathtools}

\usepackage[T1]{fontenc}

\usepackage[utf8]{inputenc}

\usepackage{microtype}

\usepackage{inconsolata}

\usepackage{graphicx}
\usepackage{booktabs}       %

\usepackage[]{todonotes}

\title{Non-Parametric Machine Text Detection via \\ Multi-View Gaussian Processes}

\author{Aleem Khan, Nicholas Andrews\\
  Department of Computer Science \\
  Johns Hopkins University \\
  \texttt{\{aleem,noa\}@cs.jhu.edu} \\ }

\begin{document}
\maketitle
\begin{abstract}
\input{tex/abstract}
\end{abstract}

\section{Introduction}

\input{tex/introduction}

\section{Preliminaries}

\subsection{Detection Under Adversarial Conditions Remains Difficult}
\input{tex/relatedwork}

\subsection{Problem Statement}\label{sec:problem}
\input{tex/problem}

\section{Method}

\input{tex/method.tex}\label{sec:method}

\section{Experiments}
\input{tex/experiments}\label{sec:experiments}

\section{Conclusion}\label{sec:conclusion}
\input{tex/conclusion}

\section*{Limitations}
We experiment with three views (stylistic, structural, and probabilistic) which our experiments show have complementary benefits enabling robustness to distribution shifts and evasion techniques. Having shown the value of multiple views for robust detection, a natural next step would be experiment with further feature spaces that could provide orthogonal benefits. For example, looking at how features vary within documents could perhaps help detect when editing has taken place. Even with the improved performance of our system, there are significant performance drops in the low FPR operating conditions, which are most like real world scenarios (e.g., plagiarism or cheating accusations in academic settings), more work is needed to validate these systems in such settings.
Separately, our experiments primarily employ existing established benchmarks for machine-text detection, which are primarily English. Our experiments in far-OOD detection suggest that our approach is robust to different languages from those the detector is fit on (appropriately raising uncertainty), but future work should validate the proposed approach in truly multilingual settings. 

\bibliography{references}

\appendix

\input{tex/appendix}

\label{sec:appendix}

\end{document}

%% file: tex/abstract.tex
  Adversarial conditions such as paraphrasing and targeted style transfer
sharply degrade the accuracy of machine text detectors.
A document, however, carries multiple complementary signals (e.g., stylistic features, likelihood and rank-order features, and structural features),
and an attack that suppresses one
may leave others intact.
While a parametric classifier can learn to combine these features given sufficient supervision,
classifiers are prone to making confidently incorrect predictions
when the distribution shifts
(e.g., novel attacks or unseen language models).
  To address this, we propose a multi-view, non-parametric detection framework that extracts complementary feature views from the same document and aggregates per-view evidence through a Gaussian process ensemble.
  By aggregating evidence across views, an adversary must simultaneously defeat multiple independent axes of detection, substantially raising the cost of evasion.
  The Gaussian process formulation additionally provides calibrated probabilities and principled abstention on out-of-distribution inputs, supporting reliable deployment in high-stakes settings.
  We evaluate on three benchmarks spanning diverse generators and attacks: the DetectRL and RAID benchmarks, and the PAN~2025 shared task and demonstrate that our multi-view detector maintains strong performance under the considered attacks, outperforming existing approaches against held out attacks.

%% file: tex/introduction.tex
As language models (LMs) have become more capable and widely available to users, text generated by LMs has become more ubiquitous and indistinguishable from human writing \citep{comanici2025gemini25pushingfrontier, grattafiori2024llama3herdmodels, openai2024gpt4technicalreport}. While LMs serve many positive use cases, and have become closely intertwined with workflows, detection of machine-generated content, especially in high-stakes domains, is increasingly of interest to many communities \cite{ippolito-etal-2020-automatic, gehring2025assessingllmtextdetection}. As generators have improved, detectors have as well: a growing body of work on machine-generated text detection, yielding zero-shot statistical tests \citep{bao2024fastdetectgpt, gehrmann-etal-2019-gltr, hans2024spotting}, trained classifiers \cite{li-etal-2024-mage, lee2024remodetect, tian2024multiscale, hu2023radar}, and commercial detection services \cite{emi2024technicalreportpangramaigenerated}. Under controlled conditions, where machine text is generated without modification or adversarial intent, these detectors achieve high accuracy \cite{hans2024spotting}. However, real-world deployment introduces a fundamentally harder problem: adversarial conditions in which machine text is \emph{edited, rewritten, or obfuscated}, with a human or another LM, before it reaches the detector \cite{thai2026editlens}.

Adversarial manipulation comes in many forms. In relatively simpler cases, a user may attempt to alter machine written documents with a prompt \citep{patel2024lowresourceauthorshipstyletransfer}. On the other end of the spectrum, a more sophisticated adversary may fine-tune generators to target specific types of detector directly \citep{nicks2024language}, or through proxies \citep{wang2025humanizing, soto2025languagemodelsoptimizedfool}. Alternatively, machine text may be passed through a trained paraphraser to disrupt rank order scores of tokens \cite{krishna2023paraphrasing}, or a pipeline of multiple paraphrasers, which can amplify this degradation. These attacks exploit different vulnerabilities, but most existing detectors rely on a single feature space, such as token-level probability under a reference language model \cite{gehrmann-etal-2019-gltr}, or stylistic fingerprints \cite{soto2024fewshot}, and a targeted edit along that axis is sufficient to evade detection. To address the multi-objective nature of detection, we propose a multi-view, non-parametric framework that exploits this insight.

 We begin by building a few-shot support for a target domain. Our method relies on having a small number of human and machine exemplars from a domain of interest\footnote{Crucially, the attacks themselves are held out in our evaluations.}. For each of $K$ views (\S\ref{sec:views}), we fit independent Gaussian process classifiers, yielding probabilities that naturally incorporate the GP's predictive uncertainty (\S\ref{sec:gp}). These probabilities are aggregated via a secondary linear model which produces a final calibrated uncertainty (\S\ref{sec:aggregation}).
 
Our contributions are as follows: \textbf{(1)} A multi-view detection framework that aggregates complementary views for robust detection under human editing and paraphrase attacks. \textbf{(2)} A Gaussian process ensemble that delivers calibrated uncertainty, and thorough analysis demonstrating the approach's robustness to various attacks. \textbf{(3)} Evaluation on diverse benchmarks: DetectRL and RAID benchmarks, and PAN 2025 shared task datasets, demonstrating strong performance under adversarial conditions where single-view detectors fail.

%% file: tex/relatedwork.tex
Detection of AI-generated and AI-manipulated content has received significant attention from the research community as generators have become more capable and accessible. Zero-shot detection methods score documents using a reference model, using the idea that machine-generated text will generally be more likely under any language model \cite{mitchell2023detectgptzeroshotmachinegeneratedtext, bao2024fastdetectgpt, hans2024spotting, su2023detectllm, yang2024dnagpt}. Parametric methods have also demonstrated strong performance, but struggle to adapt to new distributions \cite{solaiman2019release, li-etal-2024-mage, hu2023radar}. Watermarking has emerged as another effective approach for detection, but it assumes access to the model during inference \cite{pmlr-v202-kirchenbauer23a}.

Recent work has also demonstrated that many detection approaches have significant vulnerabilities to a range of attacks and adversarial conditions and we replicate these findings \cite{soto2025languagemodelsoptimizedfool, nicks2024language, krishna2023paraphrasing}. \citet{nicks2024language} in particular highlights a key risk of new detection methods, namely that they become targets to optimize against. \citet{sadasivan2025can} demonstrated that iteratively applying paraphrasing attacks significantly degrades performance.
Recently released datasets have pivoted away from evaluating on strictly machine-generated text, and consider cases where humans and LMs may edit and manipulate each others writing \cite{he2024mgtbenchbenchmarkingmachinegeneratedtext, artemova-etal-2025-beemo, dugan-etal-2024-raid, wu2024detectrl}. 
Prior work has demonstrated that Gaussian Processes trained on feature extractors are effective classifiers for synthetic speech detection \cite{glazer2025fewshotspeechdeepfakedetection}. Most closeley related to our work, Ghostbuster combines several LM derived features for detection, applies a search procedure, and applies a logistic regression to arrive at a final answer \cite{verma-etal-2024-ghostbuster}. Our approach differs in that combine distinct feature spaces, and use Gaussian processes which yield calibrated uncertainties with orders of magnitude less data.

%% file: tex/problem.tex
\begin{figure*}[t]
\includegraphics[width=\linewidth]{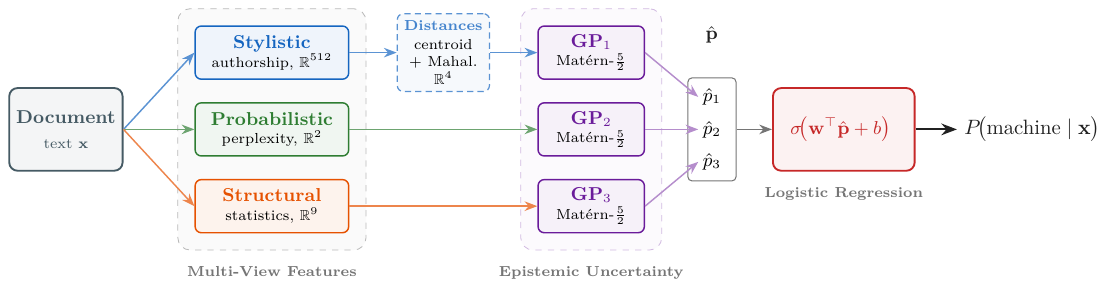}
\caption{Overview of the proposed approach. Each input $x$ is represented via three complementary views and transformed into a probability via view-specific GPs. The three scalars are then projected into a single probability via logistic regression.}
\label{fig:method}
\end{figure*}

Let $\mathcal{X}$ denote the space of natural-language documents. We define our
\emph{detector} as a function $h : \mathcal{X} \to \{0, 1\}$ that maps a
document to a binary label, where $y=0$ indicates human-written text and $y=1$
indicates machine-generated or machine-manipulated text.

\paragraph{Zero-shot detection.}
Most existing detectors operate in a \emph{zero-shot} setting: the detector is
applied directly to a test document with no domain-specific training data.
Zero-shot methods, such as log-rank tests, likelihood curvature estimates, and
cross-perplexity ratios, can easily be run on any
generator's output, in any domain, under any attack, with no adaptation.
However, this generality comes at a cost, with performance degrading substantially when the test distribution involves adversarial manipulation.%

\paragraph{Our setting: few-shot, domain-anchored detection.}
We adopt a different operating assumption. We assume our system has access
to a small \emph{support set} of in-domain documents:

\begin{equation}
    \begin{gathered}
        \mathcal{S} = \{(x_i, y_i)\}_{i=1}^{N}, \qquad y_i \in \{0,1\}, \\
        N = N_H + N_M.
    \end{gathered}
\end{equation}

where $N_H$ human-written documents and $N_M$ machine-generated (and manipulated) documents are
drawn from the domain of interest. In our main experiments we use
$N_H = N_M = 32$ documents per class to represent a quantity that is realistic to obtain
in most practical scenarios (e.g., a set of known human written essays, verified news
articles, or authentic forum posts). We also consider system performance as $N$ increases to 256 (\autoref{fig:raid-stacking-num-train}).

Crucially, we impose no requirement that the support set reflect the
\emph{attack} or the \emph{generator} encountered at test time. This generally aligns with many real-world scenarios, where anticipating all possible generators or attacks isn't feasible. The human
examples must come from the target domain, but the machine examples may be
generated by any available model, even one different from the adversary's
generator. In our evaluations, we explicitly hold out both the attack type and the source
generator from training, evaluating in a cross-attack and cross-generator
transfer setting. That is, we train on a collection of attacks and generators \emph{except} one, and evaluate on that setting. Our key finding is that the GP-based multi-view ensemble is
able to leverage this small, potentially mismatched support to generalize
robustly to held-out generators and unseen attacks
(\S\ref{sec:experiments}).

\paragraph{Why few-shot?}
This formulation takes a middle ground between the zero-shot paradigm (no
domain data, broad but fragile coverage) and fully supervised approaches (large
labeled corpora, strong but narrow). By anchoring the detector to a small
sample from the deployment domain, we provide the model with enough
distributional context to learn meaningful decision boundaries, particularly in
the centroid based feature space (\S\ref{sec:distance}), while keeping the data
requirement low enough for practical adoption.

%% file: tex/method.tex
In order to develop a robust detector, we hypothesize that learning to combine the outputs of different models fit on complementary feature spaces will have several advantages over a joint model that learns arbitrary combinations of features across views. First of all, by using independent view-specific classifiers, an adversary must defeat all views simultaneously to fully evade detection. Second, since we must learn detectors on the basis of small samples of confirmed real and fake data, a model that enabled arbitrary feature combinations would be prone to overfitting those features and therefore generalize poorly.

To this end, we present a multi-view, non-parametric framework for machine-generated text
detection. Given a document $x$, the system (i) extracts features from $K$
independent views $\{\varphi_k\}_{k=1}^K$, (ii) projects each view into a
low-dimensional \emph{distance feature} space, (iii) fits an independent
variational Gaussian process (GP) classifier per view, (iv) obtains per-view
Bernoulli probabilities $p_k$ via the probit link, (v) aggregates these
probabilities, and (vi) calibrates a decision threshold with finite-sample
false-positive guarantees. An out-of-distribution (OOD) gate enables principled
abstention when the model encounters text outside its training support.

\subsection{Multi-View Feature Extraction}
\label{sec:views}

Our approach to selecting views is simple: force an attacker to solve a multi-objective problem due to the different feature spaces of the views. As illustrated in 
While we use a simple collection of three views, we find that adding additional views helps. Each view $\varphi_k : \mathcal{X} \to \mathbb{R}^{D_k}$ maps a raw text
document to a feature vector that reflects a distinct axis of variation between
human and machine writing. We use $K=3$ views:

\paragraph{Style view ($D_k = 4$).}
Dense stylometric embeddings produced by a style representation
model\footnote{\url{https://huggingface.co/rrivera1849/LUAR-CRUD}}, which is trained to encode
writing style invariant to topic \cite{rivera-soto-etal-2021-learning}. These embeddings capture lexical and
syntactic fingerprints that are largely orthogonal to semantic content.
To avoid challenges associated with high-dimensional features, we fit centroids based on human and machine data, and compute distances from each centroid for each test point.

\paragraph{Probabilistic view ($D_k = 2$).}
A vector of zero-shot detector scores: (1) the LogRank score, which measures
the average token rank under a reference language model, and (2) the
FastDetectGPT score, which estimates the
log-likelihood curvature around the document. Both scores are computed using
Falcon-7B \cite{almazrouei2023falconseriesopenlanguage}.

\paragraph{Structural view ($D_k = 8$).}
A vector of hand-crafted document statistics which capture surface-level regularities in document organization that persist through many paraphrase operations, such as total token count and sentence count. Our exact implementation can be found in \autoref{sec:structural_view}.

The key motivation for multi-view combination is adversarial robustness.
A paraphrase attack that normalizes token-level probability features (the
probabilistic view) leaves structural regularities and stylistic fingerprints
largely intact. Requiring converging evidence
across views therefore raises the cost of any single-axis attack.

\subsection{Centroid-Based Feature Construction}
\label{sec:distance}

For high-dimensional embeddings (e.g., $D{=}512$), Euclidean distance concentration causes the GP's kernel to lose discriminative power: pairwise distances converge to a common value, and the GP posterior degenerates toward the prior. 
A simple solution is to project such views to a
 low-dimensional space. In more detail, given a labeled support set $\mathcal{S} = \{(x_i, y_i)\}_{i=1}^{N}$ (where
$y_i \in \{0, 1\}$ denotes human/machine), for each view $k$ we first compute class centroids:

\begin{equation*}
    \begin{aligned}
        \boldsymbol{\mu}_{H,k} &= \frac{1}{|\mathcal{S}_H|} \sum_{i: y_i = 0} \varphi_k(x_i) \\
        \boldsymbol{\mu}_{M,k} &= \frac{1}{|\mathcal{S}_M|} \sum_{i: y_i = 1} \varphi_k(x_i)
    \end{aligned}
\end{equation*}
where $\mathcal{S}_H$ and $\mathcal{S}_M$ denote the human and machine subsets
of the support set, respectively.
Each document is mapped to a 4D feature vector combining Euclidean and diagonal
Mahalanobis distances to both class centroids:
\begin{equation*}
    \psi_k(x) = \bigl[\,d_{H,k},\; d_{M,k},\; m_{H,k},\; m_{M,k}\,\bigr]
    \in \mathbb{R}^4,
\end{equation*}
where $d_{H,k} = \|\varphi_k(x) - \boldsymbol{\mu}_{H,k}\|_2$ and
$d_{M,k} = \|\varphi_k(x) - \boldsymbol{\mu}_{M,k}\|_2$ are Euclidean
distances, and
$m_{H,k} = \|(\varphi_k(x) - \boldsymbol{\mu}_{H,k}) /
\boldsymbol{\sigma}_{H,k}\|_2$ and
$m_{M,k} = \|(\varphi_k(x) - \boldsymbol{\mu}_{M,k}) /
\boldsymbol{\sigma}_{M,k}\|_2$ are diagonal Mahalanobis distances, with
$\boldsymbol{\sigma}_{H,k}$ and $\boldsymbol{\sigma}_{M,k}$ denoting the
per-dimension standard deviations of the human and machine support features,
respectively. The Mahalanobis components account for per-dimension variance,
making the representation sensitive to distributional shape rather than scale
alone.
Finally, the distance feature vector is standardized (median/MAD); the exact form is given
in Appendix~\ref{sec:gp_details}. We denote the standardized features
$\tilde{\psi}_k(x)$.

\subsection{Per-View Gaussian Process Classifiers}
\label{sec:gp}

A separate Gaussian process binary classifier is fit for each view $k$ on its
(optionally distance-reduced) features $\tilde{\psi}_k$. We use a sparse
variational GP with a Bernoulli likelihood and a probit link \cite{pmlr-v38-hensman15}, a
Matérn-$5/2$ kernel and inducing
locations fixed to all $N$ training points\footnote{In our regime $N$ ranges
from $8$ to $128$ per class, so this incurs no scaling concerns.}; the
variational distribution and kernel parameters are jointly trained by
maximizing the evidence lower bound \cite{pmlr-v5-titsias09a}. Full training hyperparameters are listed in
Appendix~\ref{sec:gp_details}.

At a test point $x^*$, the GP posterior over the latent function gives a
predictive mean $\mu_k(x^*) = \mathbb{E}[f_k(\tilde{\psi}_k(x^*))]$ and
variance $\sigma^2_k(x^*) = \operatorname{Var}[f_k(\tilde{\psi}_k(x^*))]$.
The Bernoulli probability of machine generation for view $k$ is obtained via the probit link:
\begin{equation*}
    p_k(x^*) = \Phi\!\left(\frac{\mu_k(x^*)}{\sqrt{1 + \sigma^2_k(x^*)}}\right),
\end{equation*}
where $\Phi$ is the standard normal CDF. This probability naturally
incorporates the GP's predictive uncertainty: when $\sigma^2_k$ is large, the
argument to $\Phi$ is shrunk toward zero, pushing $p_k$ toward $0.5$ and
reflecting the model's lack of confidence. The per-view probabilities $p_k$ are
passed directly to the aggregation stage described next.

\subsection{Calibration}
\label{sec:aggregation}

A simple max or mean over the per-view probabilities $\{p_k(x^*)\}_{k=1}^K$
ignores systematic differences in calibration across views and is sensitive to
a single poorly calibrated GP. We instead learn an
aggregator in a data-driven manner using a small calibration set.

\paragraph{Calibration set.}
Of the $2N$ labeled documents available for a (domain, generator) pair we use
the first $N/2$ per class to fit the per-view GPs and hold out the remaining
$N/2$ per class as a calibration set $\mathcal{C}$. Because the GPs never see
$\mathcal{C}$, their predictions on it are unbiased estimates of test-time
behavior, eliminating the calibration-leakage failure mode of in-sample
stacking.
\input{tex/main_tables/table1_32}

\input{tex/main_tables/table3_32}

\paragraph{Second-stage logistic regression.}
For each $x_i \in \mathcal{C}$ we collect the per-view probability vector
$\mathbf{p}(x_i) = \bigl(p_1(x_i), \dots, p_K(x_i)\bigr) \in [0,1]^K$ and fit
an $\ell_2$-regularized logistic regression.
The learned weights expose how
much each view contributes to the final detection score and absorb
challenges that any individual GP may introduce. This pipeline can be visualized in \autoref{fig:method}. We also consider alternative aggregation strategies in \autoref{sec:alternative_aggregation}.

\subsection{Evaluation Protocol}
\label{sec:eval}
Each trained model is evaluated in a \emph{cross-attack
setting}, where the model is evaluated against attacks it has never seen.
\paragraph{Metrics.}
We report partial AUROC at a maximum
false-positive rate of $1\%$ (AUROC@1\%), which stress-tests the low-FPR
regime relevant to high-stakes applications. We observe that AUROC over all operating points becomes saturated, making it difficult to observe changes between detectors. We additionally report Brier score and
Expected Calibration Error (ECE) to measure probabilistic calibration \cite{naeini2014binaryclassifiercalibrationnonparametric}, lower is better for both of these metrics.

%% file: tex/main_tables/table1_32.tex
\begin{table*}[t]
  \centering
  \caption{Detection and calibration performance on the PAN2025 benchmark at $N{=}32$ training examples per class. $\pm$ indicates sample std over three samples. \textit{MC} = machine-continued;
    \textit{MP} = machine-polished. ECE and Brier are lower-is-better.}
  \label{tab:mpu-baselines-multi-n32}
  \small
  \setlength{\tabcolsep}{4pt}
  \begin{tabular}{lcccccc}
    \toprule
    & \multicolumn{3}{c}{\textit{MC}} & \multicolumn{3}{c}{\textit{MP}} \\
    \cmidrule(lr){2-4}\cmidrule(lr){5-7}
    Method
      & AUC@1\%$\uparrow$ & ECE$\downarrow$ & Brier$\downarrow$
      & AUC@1\%$\uparrow$ & ECE$\downarrow$ & Brier$\downarrow$ \\
    \midrule
    Binoculars~\cite{hans2024spotting}
      & $.657_{\pm .000}$
      & $.169_{\pm .007}$
      & $.198_{\pm .000}$
      & $.587_{\pm .000}$
      & $.092_{\pm .001}$
      & $.207_{\pm .000}$ \\
    LRR~\cite{su2023detectllm}
      & $.513_{\pm .000}$
      & $.081_{\pm .016}$
      & $.246_{\pm .002}$
      & $.504_{\pm .000}$
      & $.050_{\pm .006}$
      & $.252_{\pm .001}$ \\
    Fast-DetectGPT~\cite{bao2024fastdetectgpt}
      & $\mathbf{.684}_{\pm .000}$
      & $.089_{\pm .003}$
      & $.189_{\pm .000}$
      & $.605_{\pm .000}$
      & $.124_{\pm .005}$
      & $.207_{\pm .001}$ \\
    Log-Rank~\cite{su2023detectllm}
      & $.500_{\pm .000}$
      & $.092_{\pm .024}$
      & $.247_{\pm .001}$
      & $.503_{\pm .000}$
      & $\mathbf{.034}_{\pm .002}$
      & $.237_{\pm .000}$ \\
    MAGE~\cite{li-etal-2024-mage}
      & $.502_{\pm .000}$
      & $.232_{\pm .000}$
      & $.231_{\pm .000}$
      & $.501_{\pm .000}$
      & $.348_{\pm .000}$
      & $.347_{\pm .000}$ \\
    RADAR~\cite{hu2023radar}
      & $.547_{\pm .000}$
      & $.183_{\pm .000}$
      & $.232_{\pm .000}$
      & $.502_{\pm .000}$
      & $.317_{\pm .000}$
      & $.352_{\pm .000}$ \\
    MPU~\cite{tian2024multiscale}
      & $.530_{\pm .000}$
      & $.436_{\pm .000}$
      & $.432_{\pm .000}$
      & $.591_{\pm .000}$
      & $.363_{\pm .000}$
      & $.360_{\pm .000}$ \\
    ReMoDetect~\cite{lee2024remodetect}
      & $.509_{\pm .000}$
      & $.172_{\pm .004}$
      & $.271_{\pm .002}$
      & $.580_{\pm .000}$
      & $.074_{\pm .006}$
      & $.177_{\pm .001}$ \\
    RoBERTa-FT
      & $.539_{\pm .046}$
      & $.117_{\pm .052}$
      & $.231_{\pm .039}$
      & $.562_{\pm .061}$
      & $.080_{\pm .033}$
      & $.214_{\pm .027}$ \\
    \midrule
    \textit{Ours}
      & $.612_{\pm .009}$
      & $\mathbf{.080}_{\pm .016}$
      & $\mathbf{.161}_{\pm .017}$
      & $\mathbf{.610}_{\pm .028}$
      & $.096_{\pm .011}$
      & $\mathbf{.177}_{\pm .007}$ \\
    \bottomrule
  \end{tabular}
\end{table*}

%% file: tex/main_tables/table3_32.tex
\begin{table*}[t]
  \centering
  \caption{Detection and calibration performance on the RAID benchmark at $N=32$ training examples per class. $\pm$ indicates sample std over three samples.}
  \label{tab:raid-gpt4-baselines-multi-n32}
  \small
  \setlength{\tabcolsep}{4pt}
  \begin{tabular}{lcccccc}
    \toprule
    & \multicolumn{3}{c}{\textit{News}} & \multicolumn{3}{c}{\textit{Reddit}} \\
    \cmidrule(lr){2-4}\cmidrule(lr){5-7}
    Method
      & AUC@1\%$\uparrow$ & ECE$\downarrow$ & Brier$\downarrow$
      & AUC@1\%$\uparrow$ & ECE$\downarrow$ & Brier$\downarrow$ \\
    \midrule
    Binoculars~\cite{hans2024spotting}
      & $.683_{\pm .124}$
      & $.255_{\pm .052}$
      & $.183_{\pm .045}$
      & $.572_{\pm .050}$
      & $.223_{\pm .041}$
      & $.182_{\pm .041}$ \\
    LRR~\cite{su2023detectllm}
      & $.513_{\pm .020}$
      & $.137_{\pm .089}$
      & $.243_{\pm .045}$
      & $.510_{\pm .008}$
      & $.152_{\pm .052}$
      & $.198_{\pm .040}$ \\
    Fast-DetectGPT~\cite{bao2024fastdetectgpt}
      & $.741_{\pm .129}$
      & $\mathbf{.105}_{\pm .100}$
      & $.113_{\pm .110}$
      & $.649_{\pm .115}$
      & $\mathbf{.095}_{\pm .095}$
      & $.135_{\pm .093}$ \\
    Log-Rank~\cite{su2023detectllm}
      & $.574_{\pm .157}$
      & $.175_{\pm .144}$
      & $.249_{\pm .094}$
      & $.527_{\pm .062}$
      & $.152_{\pm .057}$
      & $.168_{\pm .069}$ \\
    MAGE~\cite{li-etal-2024-mage}
      & $.529_{\pm .019}$
      & $.298_{\pm .051}$
      & $.298_{\pm .051}$
      & $.582_{\pm .055}$
      & $.229_{\pm .070}$
      & $.229_{\pm .070}$ \\
    RADAR~\cite{hu2023radar}
      & $.925_{\pm .057}$
      & $.232_{\pm .068}$
      & $.149_{\pm .069}$
      & $.529_{\pm .018}$
      & $.178_{\pm .078}$
      & $.172_{\pm .058}$ \\
    MPU~\cite{tian2024multiscale}
      & $.503_{\pm .015}$
      & $.429_{\pm .110}$
      & $.422_{\pm .130}$
      & $.527_{\pm .046}$
      & $.452_{\pm .078}$
      & $.432_{\pm .116}$ \\
    ReMoDetect~\cite{lee2024remodetect}
      & $.577_{\pm .069}$
      & $.213_{\pm .056}$
      & $.273_{\pm .010}$
      & $.687_{\pm .128}$
      & $.259_{\pm .105}$
      & $.167_{\pm .102}$ \\
    RoBERTa-FT
      & $.925_{\pm .157}$
      & $.143_{\pm .096}$
      & $.056_{\pm .101}$
      & $.711_{\pm .202}$
      & $.216_{\pm .066}$
      & $.091_{\pm .057}$ \\
    \midrule
    \textit{Ours}
      & $\mathbf{.930}_{\pm .104}$
      & $.235_{\pm .040}$
      & $\mathbf{.052}_{\pm .028}$
      & $\mathbf{.775}_{\pm .136}$
      & $.243_{\pm .035}$
      & $\mathbf{.088}_{\pm .027}$ \\
    \bottomrule
  \end{tabular}
\end{table*}

%% file: tex/experiments.tex
\input{tex/main_tables/table2_32}
\subsection{Datasets}
We evaluate our approach on a diverse set of datasets, each consisting of several attacks, domains, and language models\footnote{We plan to publish our splits alongside our code to reproduce results}.
\textbf{PAN2025 \cite{10.1007/978-3-032-04354-2_21}} The PAN2025 shared task considers five mixtures of human and machine writing: machine-written then human-edited, deeply-mixed text, human-initiated then machine-continued, human-written then machine-polished, and machine-written then machine-humanized\footnote{\url{https://pan.webis.de/clef25/pan25-web/generated-content-analysis.html}}. We focus our evaluations on the held out ``human-initiated then machine-continued'' (MC) and ``human-written then machine-polished'' (MP) conditions, using the remaining machine splits to form a machine support. 

\textbf{RAID \cite{dugan-etal-2024-raid}} We build custom splits from the publicly released RAID dataset to match our problem statement\footnote{\url{https://huggingface.co/datasets/liamdugan/raid}}. Specifically, we sample data from three diverse domains (News, Reddit, and Amazon Reviews) and consider all attacks in the dataset \cite{dugan-etal-2024-raid}. Within each domain, similar to DetectRL, to form a few-shot machine support, we sample exemplars from all attacks except one evaluation attack, the evaluation split contains human data, and machine sample from that attack only. We also control for language model: GPT-4o is held out from all training splits and only included as a held out model during evaluation.

\textbf{DetectRL \cite{wu2024detectrl}} We use the exact splits released by the authors to evaluate our detector\footnote{\url{https://github.com/NLP2CT/DetectRL}}. We evaluate on the ``Multi-Attack'' setting in our main experiments to match the problem statement outlined in \autoref{sec:problem} to control for domain. To form the few-shot machine support, we sample exemplars from all attacks \emph{except} for one evaluation attack, the evaluation split contains human data, and machine samples from that attack only.%

\subsection{Baselines}\label{sec:baselines}
We compare our approach to several popular zero-shot methods: Binoculars \cite{hans2024spotting}, Fast-DetectGPT \cite{bao2024fastdetectgpt}, Log-Likelihood Log-Rank Ratio (LRR) \cite{su2023detectllm}, Log-Rank \cite{su2023detectllm}, MAGE \cite{li-etal-2024-mage}, RADAR \cite{hu2023radar}, ReMoDetect \cite{lee2024remodetect}, and MPU \cite{tian2024multiscale}. Some of these approaches are classifier based, and return interpretable scores between zero and one to discriminate between human and machine text, while others return a raw score. To compute calibration metrics on the latter, we calibrate the raw scores using Platt scaling on held out calibration data \cite{Platt1999ProbabilisticOF}. Noting that the previous approaches are all zero-shot and do not make use of the data our system uses beyond calibration, we also train a transformer based binary classifier on the same available to our system by fine-tuning RoBERTa \cite{liu2019robertarobustlyoptimizedbert}. Details for this classifier can be found in \autoref{sec:base_classifier_details}.

\subsection{Main Experiments} \label{sec:main_experiments}

\begin{figure*}[ht]
  \centering
  \includegraphics[width=\linewidth]{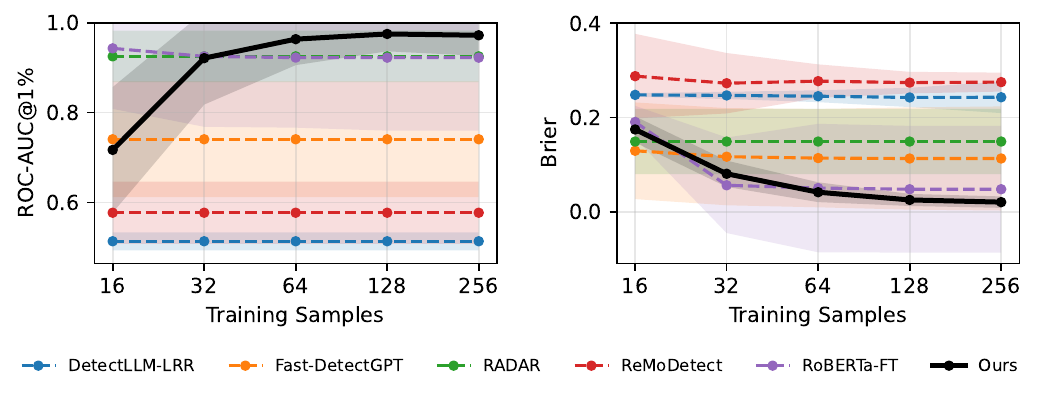}
  \caption{Detection (left) and calibration (right) performance on the News split of the RAID dataset. The GP based approach is able to quickly reach a strong performance point with a small number of training samples, variance of results also decreases significantly with more training samples. Shaded regions show $\pm$1 standard error of the mean across different samples and evaluation conditions.}
  \label{fig:raid-stacking-num-train}
\end{figure*}

This section describes our experiments across adversarial data splits from the PAN2025 shared task \autoref{tab:mpu-baselines-multi-n32}, RAID benchmark \autoref{tab:raid-gpt4-baselines-multi-n32}, and the DetectRL benchmark \autoref{tab:detectrl-attacks-multi-n32}. Due to space constraints, we report results for two domains from the RAID dataset in \autoref{tab:raid-gpt4-baselines-multi-n32} and for two attacks from the DetectRL benchmark in \autoref{tab:detectrl-attacks-multi-n32}. Additional results with similar trends can be found in \autoref{sec:addtional_domains_attacks}. Firstly, we confirm findings from previous work and demonstrate that state-of-the-art detectors are brittle to attacks \cite{krishna2023paraphrasing, soto2025languagemodelsoptimizedfool}, especially under low FPR operating conditions. The brittleness of these systems extends to calibration, further analysis of this consequence can be found in the following section. Throughout our results, the fine-tuned RoBERTa classifier outperforms most baselines and is most competitive with our system. We do note that the error bars for this baseline are significantly wider for many conditions, indicating that training is sensitive to the few-shot sample selected. Beyond RoBERTa, across most conditions considered, our few-shot approach significantly outperforms the considered zero-shot baselines under both detection, and calibration metrics.

We report test results on all three datasets, with hyperparameters for our system chosen based on a separate development split of the PAN2025 dataset. \autoref{tab:mpu-baselines-multi-n32} shows performance under two difficult machine splits: human-initiated and machine-continued (MC) and human-written and machine-polished (MP). These results have the lowest absolute scores relative to other datasets due to the level of mixing human and machine text. \autoref{tab:raid-gpt4-baselines-multi-n32} shows performance for the three considered RAID domains, with performance averaged across all 11 considered attacks. Full results for each attack can be found in \autoref{sec:extra_raid}. %

\textbf{How does system performance scale with few-shot samples?} The baselines we consider are zero-shot, however we do make use of few-shot samples to calibrate these approaches (\S\ref{sec:baselines}). \autoref{fig:raid-stacking-num-train} shows system performance against number of few-shot training samples on the News split of the RAID dataset. All approaches are calibrated on the same data, and results are averaged across all 11 attacks. Our approach is able to make use of small amounts of data for detection and there is significantly less variance associated with the selection of this data compared to other methods. Notably, system calibration improves significantly with the number of samples.

\begin{figure}
    \centering
    \includegraphics[width=\columnwidth]{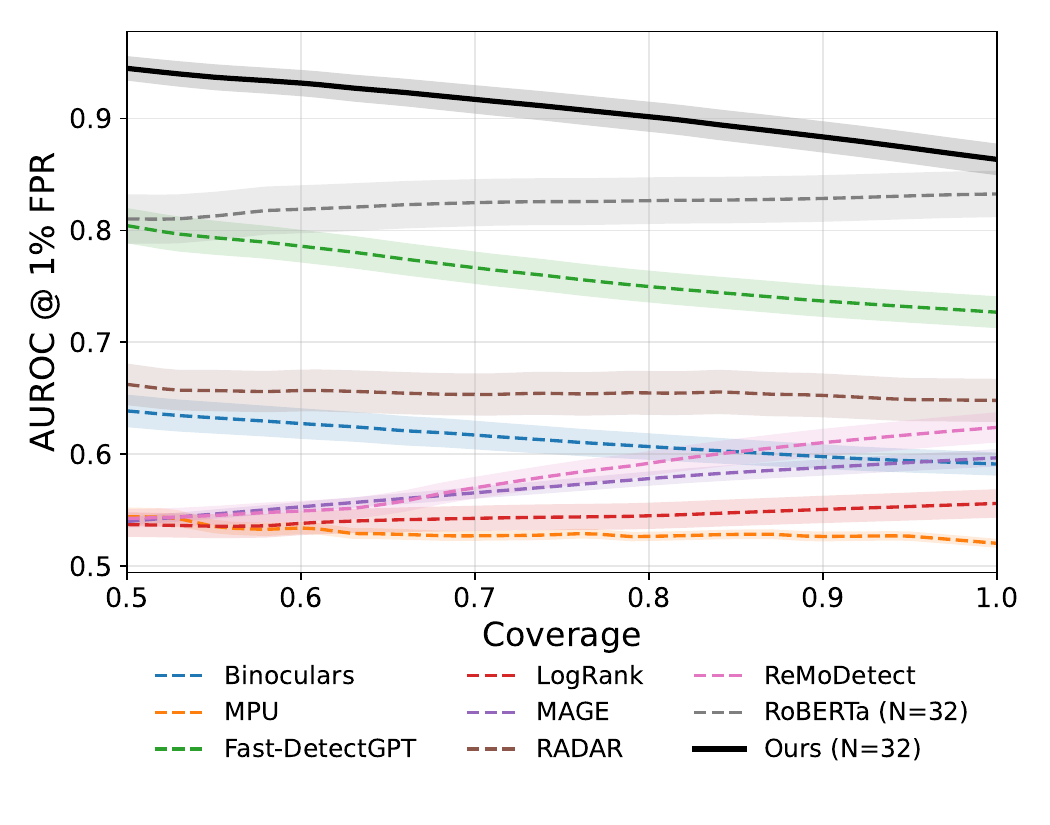}
    \caption{Performance-Coverage curves on the News split of the RAID dataset. Our system demonstrates a monotonically decreasing trend as it is forced to evaluate samples it is increasingly uncertain about.}
    \label{fig:abstention_paraphrase_roc_auc}
\end{figure}

\textbf{How effectively does our system trade dataset coverage for performance?} We visualize this trade-off using a Performance-Coverage curve, an adaptation of the standard Risk-Coverage formulation \cite{10.5555/3295222.3295241}. \autoref{fig:abstention_paraphrase_roc_auc} shows performance (AUROC@1\%) against dataset coverage on test data consisting of a held out a paraphrasing attack. To generate these curves we rank all test data by each system's uncertainty, as more of dataset is covered (x-axis), the system is forced to produce responses it is more and more uncertain about. Across the three domains in the RAID dataset we observe that the GP's uncertainty is indeed informative, as abstention decreases, performance monotonically decreases. We observe that several baselines produce the opposite result with performance increasing as abstention decreases, further confirming poor calibration.

\begin{figure}
    \centering
    \includegraphics[width=\columnwidth]{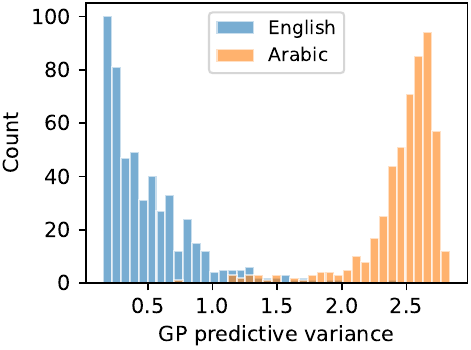}
    \caption{Predictive variance of a GP fit on a style view. English news-wire data is observed during training, at test time far-OOD Arabic data results in significantly higher variances raising the uncertainty of the system.}
    \label{fig:gp-ood-style-histograms}
\end{figure}

\textbf{How does the system handle far OOD data?} To further understand how well our system handles uncertainty we consider far-OOD data. Our main results feature held out, OOD attacks, but test data is from the same domain as the few-shot exemplars simulating a near-OOD setting. A well calibrated system should be uncertain about predictions on far-OOD data, rather than confidently incorrect (e.g., false positives when a detector observes human-written text in a previously unobserved language). To simulate this setting, we introduce human-written and machine-generated Arabic news-wire data from the M4 dataset \cite{wang-etal-2024-m4}. Figure \ref{fig:gp-ood-style-histograms} demonstrates how a GP capturing our style view handles the far-OOD data compared to near-OOD data (english data from the News split of the RAID dataset). The posterior variance significantly increases for the Arabic data, which differs stylistically from any of the training data (human or machine).

\textbf{How do different views affect system performance?}
\autoref{tab:raid-view-ablation} shows an ablation on view configurations. We find that combining views improves detection performance across all views.
\begin{table}[t]
  \centering
  \caption{Per-view ablation on the RAID dataset. Best per column in \textbf{bold}.}
  \label{tab:raid-view-ablation}
  \small
  \setlength{\tabcolsep}{4pt}
  \begin{tabular}{lcccc}
    \toprule
    Views & AUC & @1\% & ECE$\downarrow$ & Brier$\downarrow$ \\
    \midrule
    Probabilistic                    & 0.932 & 0.773 & 0.101 & 0.098 \\
    Structural                      & 0.907 & 0.663 & 0.203 & 0.100 \\
    Style                           & 0.974 & 0.685 & 0.116 & 0.069 \\
    All three          & \textbf{0.996} & \textbf{0.884} & \textbf{0.083} & \textbf{0.030} \\
    \bottomrule
  \end{tabular}
\end{table}

%% file: tex/main_tables/table2_32.tex
\begin{table*}
  \centering
  \caption{Detection and calibration performance on the DetectRL benchmark at $N=32$ training examples per class. $\pm$ indicates sample std over three samples.}
  \label{tab:detectrl-attacks-multi-n32}
  \small
  \setlength{\tabcolsep}{4pt}
  \begin{tabular}{lcccccc}
    \toprule
    & \multicolumn{3}{c}{\textit{Paraphrase}}
    & \multicolumn{3}{c}{\textit{Perturbation}} \\
    \cmidrule(lr){2-4}\cmidrule(lr){5-7}
    Method
      & AUC@1\%$\uparrow$ & ECE$\downarrow$ & Brier$\downarrow$
      & AUC@1\%$\uparrow$ & ECE$\downarrow$ & Brier$\downarrow$ \\
    \midrule
    Binoculars~\cite{hans2024spotting}
      & $.695_{\pm .000}$
      & $.151_{\pm .010}$
      & $.180_{\pm .003}$
      & $.572_{\pm .000}$
      & $.147_{\pm .004}$
      & $.212_{\pm .001}$ \\
    LRR~\cite{su2023detectllm}
      & $.616_{\pm .000}$
      & $.135_{\pm .006}$
      & $.229_{\pm .001}$
      & $.510_{\pm .000}$
      & $.038_{\pm .014}$
      & $.250_{\pm .001}$ \\
    Fast-DetectGPT~\cite{bao2024fastdetectgpt}
      & $.627_{\pm .000}$
      & $.074_{\pm .009}$
      & $.172_{\pm .001}$
      & $.555_{\pm .000}$
      & $.104_{\pm .002}$
      & $.196_{\pm .001}$ \\
    Log-Rank~\cite{su2023detectllm}
      & $.587_{\pm .000}$
      & $.095_{\pm .046}$
      & $.244_{\pm .001}$
      & $.500_{\pm .000}$
      & $\mathbf{.036}_{\pm .007}$
      & $.244_{\pm .000}$ \\
    MAGE~\cite{li-etal-2024-mage}
      & $.699_{\pm .000}$
      & $.158_{\pm .000}$
      & $.158_{\pm .000}$
      & $.561_{\pm .000}$
      & $.261_{\pm .000}$
      & $.260_{\pm .000}$ \\
    RADAR~\cite{hu2023radar}
      & $.706_{\pm .000}$
      & $.141_{\pm .000}$
      & $.135_{\pm .000}$
      & $.634_{\pm .000}$
      & $.194_{\pm .000}$
      & $.178_{\pm .000}$ \\
    MPU~\cite{tian2024multiscale}
      & $.632_{\pm .000}$
      & $.262_{\pm .000}$
      & $.273_{\pm .000}$
      & $.521_{\pm .000}$
      & $.402_{\pm .000}$
      & $.409_{\pm .000}$ \\
    ReMoDetect~\cite{lee2024remodetect}
      & $.754_{\pm .000}$
      & $\mathbf{.066}_{\pm .001}$
      & $.141_{\pm .000}$
      & $.634_{\pm .000}$
      & $.039_{\pm .006}$
      & $.139_{\pm .001}$ \\
    RoBERTa-FT Classifier
      & $.635_{\pm .061}$
      & $.188_{\pm .051}$
      & $.161_{\pm .023}$
      & $.603_{\pm .042}$
      & $.104_{\pm .043}$
      & $.176_{\pm .032}$ \\
    \midrule
    \textit{Ours}
      & $\mathbf{.804}_{\pm .020}$
      & $.137_{\pm .020}$
      & $\mathbf{.086}_{\pm .005}$
      & $\mathbf{.758}_{\pm .038}$
      & $.124_{\pm .014}$
      & $\mathbf{.114}_{\pm .024}$ \\
    \bottomrule
  \end{tabular}
\end{table*}

%% file: tex/conclusion.tex
Reliable detection of generated texts is complicated 
by the diversity of text genres, language models, prompting strategies, 
and evasion techniques.
A universal detector covering all conceivable settings may always be vulnerable to a new evasion strategy.
This paper addresses a more modest goal: how to developed specialized detectors for specific settings, and how to characterize the data that those detectors will work well on via calibrated uncertainty estimates.
To this end, we have shown that the proposed non-parametric approach enables reliable detection using small amounts of human and machine-generated data from the target distribution, that it yields well-calibrated uncertainty estimates, and that the uncertainty estimates can detect both near- and far-OOD data. Our experiments confirm that this is due to two key design choices: (1) the use of multiple complementary views capturing different facets of generated text; (2) the GP classification framework built on top of the distinct feature spaces and then aggregated via a learned calibration map.

%% file: tex/appendix.tex
\section{Gaussian Process Training Details}\label{sec:gp_details}

\subsection{Robust Standardization of Distance Features}
\label{app:robust_std}

Before being passed to the per-view GP kernels, the distance features
$\psi_k(x)$ defined in Section~\ref{sec:distance} are normalized to a
dimensionless scale using the support set's median and median absolute
deviation (MAD):
\begin{equation}
    \tilde{\psi}_k(x) = \frac{\psi_k(x) - \operatorname{median}[\psi_k(\mathcal{S})]}
                             {\operatorname{MAD}[\psi_k(\mathcal{S})] \times 1.4826},
\end{equation}
where the factor $1.4826$ makes the MAD consistent with the standard deviation
under Gaussian data. The use of median/MAD rather than mean/standard deviation
makes the standardization robust to the heavy-tailed distance distributions
that arise under adversarial attacks. This maps all views to a common scale,
making GP kernel hyperparameters and predictive probabilities directly
comparable across views.

\subsection{Training Hyperparameters}

Table~\ref{tab:gp_hyperparams} lists the hyperparameters used to train each
per-view GP. The same settings are used for every view, dataset, and
training-set size in our experiments. All experiments were conducted on a single V100 GPU.

\begin{table}[t]
\centering
\small
\begin{tabular}{ll}
\toprule
Hyperparameter & Value \\
\midrule
Kernel                       & Matérn-$5/2$ \\
Likelihood                   & Bernoulli (probit link) \\
Training Samples              & $\leq256$ \\
Variational distribution     & Cholesky \\
Optimizer                    & Adam \\
Learning rate                & $0.01$ \\
Max epochs                   & $500$ \\
Early stopping patience      & $20$ epochs \\
Early stopping min.\ delta   & $10^{-4}$ \\
\bottomrule
\end{tabular}
\caption{Per-view GP training hyperparameters.}
\label{tab:gp_hyperparams}
\end{table}

\subsection{Structural View Implementation}\label{sec:structural_view}

The structural view summarizes each document with eight surface-level
features that are cheap to compute and require no learned model.
Tokenization uses NLTK's \texttt{punkt\_tab} sentence and word tokenizers;
paragraphs are approximated by splitting on double newlines.
Table~\ref{tab:structural_features} lists the features.

\begin{table}[t]
\centering
\small
\begin{tabular}{ll}
\toprule
Feature & Definition \\
\midrule
Doc length        & Number of word tokens. \\
Sentence count         & Number of sentences. \\
Paragraph count        & Number of paragraphs. \\
Mean sentence length   & Mean words per sentence. \\
Std.\ sentence length  & Population std.\ of sent. \\
Punctuation density    & Count of \{\texttt{,}, \texttt{;}, \texttt{---}, \texttt{--}\}. \\
Type-token ratio       & $|\text{unique tokens}| / |\text{tokens}|$. \\
Mean word length       & Mean chars per word. \\
\bottomrule
\end{tabular}
\caption{Per-document structural features. Each document is represented as
an $\mathbb{R}^{8}$ vector.}
\label{tab:structural_features}
\end{table}

\section{Baseline Classifier Training Details}\label{sec:base_classifier_details}

We fine-tune RoBERTa-base on the exact same data splits seen by our system \footnote{\url{https://huggingface.co/FacebookAI/roberta-base}}. We train for 10 epochs using the AdamW optimizer and a learning rate of $2e^{-5}$

\section{Alternative Aggregation Strategies}\label{sec:alternative_aggregation}
Here we explore several alternative approaches to aggregating features and probabilities across views. \autoref{tab:detectrl_cross_attack_detail} shows cross-attack performance with these strategies. First we consider eliminating the multi-classifier approach in favor of concatenated features fed to either a logistic regression or multi-layer perceptron (MLP) "LR/MLP (Concat)". This gives the system the ability to see all features at the same time, but makes single-view attacks more difficult to identify as out of distribution. We then consider replacing the multi-GP configuration with a multiple logistic regression or multiple MLP classifier with mean pooled probabilities. We find that the stacked GP configuration provides the most consistent performance.

\begin{table*}[t]
\centering
\small
\caption{DetectRL Cross-Attack Generalization with various aggregation strategies. \textbf{Bold} indicates best performance, \underline{underline} indicates 2nd best per column. \emph{Data Mix} attack does not contain in-domain data.}
\label{tab:detectrl_cross_attack_detail}
\setlength{\tabcolsep}{4pt}
\begin{tabular}{llcccc}
\toprule
\textbf{Eval} & \textbf{Method} & \textbf{AUCROC} & \textbf{AUROC@1} & \textbf{ECE} & \textbf{Brier} \\
\midrule
\textit{Paraphrase} & LR (Concat) & 0.9676 & 0.8327 & \textbf{0.0388} & \underline{0.0675} \\
 & LR (Per-View) & 0.9726 & 0.8326 & 0.1767 & 0.1003 \\
 & MLP (Concat) & 0.9669 & 0.8027 & 0.0539 & 0.0731 \\
 & MLP (Per-View) & \underline{0.9751} & \underline{0.8451} & 0.1527 & 0.0881 \\
 & GP (Concat) & 0.9689 & 0.7665 & 0.0817 & 0.0765 \\
 & GP (Per-View) & \underline{0.9762} & 0.8343 & 0.2233 & 0.1129 \\
 & GP (Stacking) & \textbf{0.9847} & \textbf{0.8656} & \underline{0.0529} & \textbf{0.0493} \\
\midrule
\textit{Perturbation} & LR (Concat) & 0.9703 & 0.8273 & 0.0723 & 0.0782 \\
 & LR (Per-View) & 0.9638 & 0.8067 & 0.1755 & 0.1158 \\
 & MLP (Concat) & 0.9629 & 0.7589 & \textbf{0.0686} & 0.0876 \\
 & MLP (Per-View) & \underline{0.9741} & \underline{0.8397} & 0.1647 & 0.1003 \\
 & GP (Concat) & 0.9723 & 0.8186 & 0.0777 & \underline{0.0753} \\
 & GP (Per-View) & \underline{0.9763} & \underline{0.8493} & 0.2157 & 0.1149 \\
 & GP (Stacking) & \textbf{0.9770} & \textbf{0.8618} & \underline{0.0693} & \textbf{0.0673} \\
\midrule
\textit{Prompt} & LR (Concat) & 0.9401 & \underline{0.8424} & \underline{0.0630} & 0.0921 \\
 & LR (Per-View) & 0.9453 & 0.8081 & 0.1398 & 0.1140 \\
 & MLP (Concat) & \underline{0.9510} & \textbf{0.8434} & 0.0721 & \underline{0.0894} \\
 & MLP (Per-View) & 0.9475 & 0.7951 & 0.1096 & 0.1026 \\
 & GP (Concat) & \underline{0.9521} & 0.7344 & 0.0768 & 0.0909 \\
 & GP (Per-View) & 0.9429 & 0.7731 & 0.1768 & 0.1271 \\
 & GP (Stacking) & \textbf{0.9552} & 0.8042 & \textbf{0.0444} & \textbf{0.0792} \\
\midrule
\textit{Data Mix} & LR (Concat) & 0.9372 & \textbf{0.7710} & \underline{0.0438} & 0.0976 \\
 & LR (Per-View) & 0.9376 & \underline{0.7469} & 0.1485 & 0.1248 \\
 & MLP (Concat) & 0.9415 & 0.7322 & 0.0839 & 0.1025 \\
 & MLP (Per-View) & \underline{0.9474} & 0.7212 & 0.1447 & 0.1112 \\
 & GP (Concat) & 0.9390 & 0.7017 & 0.0439 & \underline{0.0969} \\
 & GP (Per-View) & \underline{0.9483} & 0.7192 & 0.1930 & 0.1300 \\
 & GP (Stacking) & \textbf{0.9532} & 0.7331 & \textbf{0.0379} & \textbf{0.0829} \\
\bottomrule
\end{tabular}
\end{table*}

\section{Additional RAID Experiments}\label{sec:extra_raid}
\autoref{tab:raid-gpt4-baselines-multi-n32} averages results over the 11 considered attacks in the RAID benchmark. We breakdown performance across those attacks in \autoref{tab:raid-roc1pct-news}, \autoref{tab:raid-roc1pct-reddit}, and \autoref{tab:raid-roc1pct-reviews}.
\newcommand{\rot}[1]{\rotatebox{90}{\strut #1}}                               \newcommand{\ch}[1]{\textit{#1}}   %
\newcommand{\valstd}[2]{#1\,\ensuremath{{\scriptstyle \pm #2}}}

\begin{table*}
  \centering
  \caption{AUROC@1\% per attack --- \textit{News} split of RAID dataset. Italics indicate near-chance performance ($<0.55$).}
  \label{tab:raid-roc1pct-news}
  \footnotesize
  \setlength{\tabcolsep}{3pt}
  \begin{tabular}{lrrrrrrrrrrr}
    \toprule
    & \rot{AltSpell} & \rot{ArtDel} & \rot{Homoglyph} & \rot{InsPara} & \rot{Number} & \rot{Paraphrase} & \rot{PplxMisp} & \rot{Synonym} & \rot{UpLower} & \rot{Whitespace} & \rot{ZeroWidth} \\
    \midrule
    MPU~\cite{tian2024multiscale}  & \ch{0.500} & \ch{0.499} & \ch{0.497} & \ch{0.498} & \ch{0.500} & 0.551 & \ch{0.498} & \ch{0.497} & \ch{0.497} & \ch{0.497} & \ch{0.497} \\
    Binoculars~\cite{hans2024spotting}    & 0.746 & 0.611 & \ch{0.507} & 0.714 & 0.744 & 0.839 & 0.723 & \ch{0.499} & 0.636 & 0.591 & 0.904 \\
    LRR~\cite{su2023detectllm} & \ch{0.507} & \ch{0.499} & \ch{0.536} & \ch{0.506} & \ch{0.507} & \ch{0.516} & \ch{0.503} & \ch{0.498} & \ch{0.498} & \ch{0.508} & 0.567 \\
    Fast-DetectGPT~\cite{bao2024fastdetectgpt} & 0.792 & 0.649 & 0.653 & 0.763 & 0.791 & \textbf{0.886} & 0.768 & \ch{0.500} & 0.688 & 0.663 & 0.999 \\
    Log-Rank~\cite{su2023detectllm}       & \ch{0.513} & \ch{0.499} & 0.765 & \ch{0.514} & \ch{0.510} & \ch{0.512} & \ch{0.506} & \ch{0.497} & \ch{0.498} & \ch{0.498} & \textbf{1.000} \\
    MAGE~\cite{li-etal-2024-mage}          & \ch{0.535} & \ch{0.511} & \ch{0.497} & \ch{0.536} & \ch{0.538} & 0.566 & \ch{0.526} & \ch{0.513} & \ch{0.509} & \ch{0.536} & \ch{0.546} \\
    RADAR~\cite{hu2023radar}         & 0.952 & 0.921 & 0.876 & 0.964 & 0.957 & 0.777 & 0.947 & 0.920 & 0.959 & 0.907 & 0.999 \\
    ReMoDetect~\cite{lee2024remodetect}    & 0.645 & 0.554 & \ch{0.497} & 0.665 & 0.626 & \ch{0.533} & 0.643 & \ch{0.497} & \ch{0.523} & 0.665 & \ch{0.497} \\
    \midrule
    Ours          & \textbf{0.997} & \textbf{0.981} & \textbf{0.999} & \textbf{0.988} & \textbf{0.991} & 0.880 & \textbf{0.985} & \textbf{0.943} & \textbf{0.982} & \textbf{0.984} & \textbf{1.000} \\
    \bottomrule
  \end{tabular}
\end{table*}

\begin{table*}
  \centering
  \caption{AUROC@1\% per attack --- \textit{Reddit} split of RAID dataset. Italics indicate near-chance performance ($<0.55$).}
  \label{tab:raid-roc1pct-reddit}
  \footnotesize
  \setlength{\tabcolsep}{3pt}
  \begin{tabular}{lrrrrrrrrrrr}
    \toprule
    & \rot{AltSpell} & \rot{ArtDel} & \rot{Homoglyph} & \rot{InsPara} & \rot{Number} & \rot{Paraphrase} & \rot{PplxMisp} & \rot{Synonym} & \rot{UpLower} & \rot{Whitespace} & \rot{ZeroWidth} \\
    \midrule
    MPU~\cite{tian2024multiscale}  & \ch{0.504} & \ch{0.503} & 0.598 & \ch{0.503} & \ch{0.506} & \ch{0.544} & \ch{0.502} & \ch{0.502} & \ch{0.500} & \ch{0.498} & 0.637 \\
    Binoculars~\cite{hans2024spotting}    & 0.615 & 0.568 & \ch{0.498} & 0.599 & 0.632 & 0.609 & 0.595 & \ch{0.500} & \ch{0.538} & \ch{0.514} & 0.628 \\
    LRR~\cite{su2023detectllm} & \ch{0.510} & \ch{0.505} & \ch{0.510} & \ch{0.511} & \ch{0.512} & \ch{0.520} & \ch{0.508} & \ch{0.499} & \ch{0.502} & \ch{0.509} & \ch{0.531} \\
    Fast-DetectGPT~\cite{bao2024fastdetectgpt} & 0.676 & 0.607 & 0.555 & 0.658 & 0.699 & 0.702 & 0.651 & \ch{0.500} & 0.580 & 0.563 & \textbf{0.949} \\
    Log-Rank~\cite{su2023detectllm}       & \ch{0.504} & \ch{0.502} & 0.570 & \ch{0.504} & \ch{0.504} & \ch{0.504} & \ch{0.503} & \ch{0.498} & \ch{0.499} & \ch{0.500} & 0.709 \\
    MAGE~\cite{li-etal-2024-mage}          & 0.594 & \ch{0.549} & \ch{0.497} & 0.615 & 0.614 & 0.687 & 0.554 & \ch{0.535} & \ch{0.516} & 0.615 & 0.630 \\
    RADAR~\cite{hu2023radar}         & 0.550 & \ch{0.541} & \ch{0.499} & \ch{0.548} & 0.553 & \ch{0.530} & \ch{0.536} & \ch{0.527} & \ch{0.523} & \ch{0.515} & \ch{0.498} \\
    ReMoDetect~\cite{lee2024remodetect}    & 0.798 & 0.745 & \ch{0.497} & 0.814 & 0.809 & 0.651 & 0.772 & \ch{0.515} & 0.645 & 0.814 & \ch{0.497} \\
    \midrule
    Ours          & \textbf{0.901} & \textbf{0.821} & \textbf{0.859} & \textbf{0.897} & \textbf{0.961} & \textbf{0.736} & \textbf{0.894} & \textbf{0.560} & \textbf{0.848} & \textbf{0.816} & 0.633 \\
    \bottomrule
  \end{tabular}
\end{table*}

\begin{table*}
  \centering
  \caption{AUROC@1\% per attack --- \textit{Reviews} split of RAID dataset. Italics indicate near-chance performance ($<0.55$).}
  \label{tab:raid-roc1pct-reviews}
  \footnotesize
  \setlength{\tabcolsep}{3pt}
  \begin{tabular}{lrrrrrrrrrrr}
    \toprule
    & \rot{AltSpell} & \rot{ArtDel} & \rot{Homoglyph} & \rot{InsPara} & \rot{Number} & \rot{Paraphrase} & \rot{PplxMisp} & \rot{Synonym} & \rot{UpLower} & \rot{Whitespace} & \rot{ZeroWidth} \\
    \midrule
    MPU\cite{tian2024multiscale}  & \ch{0.506} & \ch{0.500} & 0.583 & \ch{0.499} & \ch{0.514} & 0.638 & \ch{0.508} & \ch{0.506} & \ch{0.498} & \ch{0.497} & 0.575 \\
    Binoculars\cite{hans2024spotting}    & \ch{0.533} & \ch{0.504} & \ch{0.497} & \ch{0.531} & 0.567 & \ch{0.512} & \ch{0.541} & \ch{0.497} & \ch{0.504} & \ch{0.499} & \ch{0.528} \\
    LRR\cite{su2023detectllm} & \ch{0.497} & \ch{0.497} & \ch{0.513} & \ch{0.497} & \ch{0.497} & \ch{0.498} & \ch{0.497} & \ch{0.497} & \ch{0.497} & \ch{0.498} & 0.595 \\
    Fast-DetectGPT\cite{bao2024fastdetectgpt} & 0.844 & 0.728 & 0.554 & 0.871 & 0.921 & \textbf{0.897} & 0.879 & \ch{0.505} & 0.771 & 0.729 & \textbf{1.000} \\
    Log-Rank\cite{su2023detectllm}       & \ch{0.504} & \ch{0.498} & \textbf{0.783} & \ch{0.505} & \ch{0.508} & \ch{0.506} & \ch{0.503} & \ch{0.497} & \ch{0.497} & \ch{0.499} & 0.928 \\
    MAGE\cite{li-etal-2024-mage}          & 0.629 & \ch{0.516} & \ch{0.497} & 0.646 & 0.645 & 0.705 & 0.626 & 0.594 & \ch{0.524} & 0.646 & 0.670 \\
    RADAR\cite{hu2023radar}         & \ch{0.497} & \ch{0.497} & \ch{0.497} & \ch{0.497} & \ch{0.497} & \ch{0.497} & \ch{0.497} & \ch{0.497} & \ch{0.497} & \ch{0.497} & \ch{0.497} \\
    ReMoDetect\cite{lee2024remodetect}    & 0.786 & 0.712 & \ch{0.497} & 0.807 & 0.804 & 0.609 & 0.793 & \ch{0.546} & 0.694 & 0.807 & \ch{0.497} \\
    \midrule
    Ours          & \textbf{0.982} & \textbf{0.905} & 0.714 & \textbf{0.984} & \textbf{0.990} & 0.788 & \textbf{0.989} & \textbf{0.601} & \textbf{0.981} & \textbf{0.951} & 0.984 \\
    \bottomrule
  \end{tabular}
\end{table*}

\section{Additional Domains and Attacks}\label{sec:addtional_domains_attacks}
Due to space limitations, \autoref{tab:raid-gpt4-baselines-multi-n32} only reported results on two domains and \autoref{tab:detectrl-attacks-multi-n32} only reported results on two attacks. We report an additional domain from the RAID dataset (\autoref{tab:raid-gpt4-reviews-n32}) and attack from the DetectRL benchmark (\autoref{tab:detectrl-prompt-n32}). The results on these extra splits share similar trends with our main experiments in \autoref{sec:main_experiments}.

\input{tex/main_tables/table3_32_reviews}

\input{tex/main_tables/table2_32_prompt}

%% file: tex/main_tables/table3_32_reviews.tex
\begin{table}
  \centering
  \caption{Detection and calibration performance on the RAID \textit{Reviews} domain at $N=32$ training examples per class. $\pm$ indicates sample std over three samples.}
  \label{tab:raid-gpt4-reviews-n32}
  \small
  \setlength{\tabcolsep}{4pt}
  \begin{tabular}{lccc}
    \toprule
    Method
      & AUC@1\%$\uparrow$ & ECE$\downarrow$ & Brier$\downarrow$ \\
    \midrule
    Binoculars~\cite{hans2024spotting}
      & $.518_{\pm .020}$
      & $.264_{\pm .046}$
      & $.176_{\pm .044}$ \\
    LRR~\cite{su2023detectllm}
      & $.508_{\pm .027}$
      & $.235_{\pm .075}$
      & $.263_{\pm .074}$ \\
    Fast-DetectGPT~\cite{bao2024fastdetectgpt}
      & $.791_{\pm .148}$
      & $\mathbf{.103}_{\pm .090}$
      & $.098_{\pm .099}$ \\
    Log-Rank~\cite{su2023detectllm}
      & $.566_{\pm .142}$
      & $.198_{\pm .106}$
      & $.226_{\pm .102}$ \\
    MAGE~\cite{li-etal-2024-mage}
      & $.609_{\pm .066}$
      & $.161_{\pm .070}$
      & $.159_{\pm .070}$ \\
    RADAR~\cite{hu2023radar}
      & $.497_{\pm .000}$
      & $.208_{\pm .100}$
      & $.171_{\pm .084}$ \\
    MPU~\cite{tian2024multiscale}
      & $.530_{\pm .046}$
      & $.415_{\pm .104}$
      & $.388_{\pm .136}$ \\
    ReMoDetect~\cite{lee2024remodetect}
      & $.687_{\pm .123}$
      & $.326_{\pm .120}$
      & $.174_{\pm .133}$ \\
    RoBERTa-FT
      & $.861_{\pm .205}$
      & $.215_{\pm .070}$
      & $.087_{\pm .053}$ \\
    \midrule
    \textit{Ours}
      & $\mathbf{.895}_{\pm .140}$
      & $.232_{\pm .037}$
      & $\mathbf{.079}_{\pm .033}$ \\
    \bottomrule
  \end{tabular}
\end{table}

%% file: tex/main_tables/table2_32_prompt.tex
\begin{table}
  \centering
  \caption{Detection and calibration performance on the DetectRL \textit{Prompt} attack at $N=32$ training examples per class. $\pm$ indicates sample std over three samples.}
  \label{tab:detectrl-prompt-n32}
  \small
  \setlength{\tabcolsep}{4pt}
  \begin{tabular}{lccc}
    \toprule
    Method
      & AUC@1\%$\uparrow$ & ECE$\downarrow$ & Brier$\downarrow$ \\
    \midrule
    Binoculars~\cite{hans2024spotting}
      & $.805_{\pm .000}$
      & $.140_{\pm .001}$
      & $.132_{\pm .000}$ \\
    LRR~\cite{su2023detectllm}
      & $.688_{\pm .000}$
      & $.091_{\pm .007}$
      & $.178_{\pm .002}$ \\
    Fast-DetectGPT~\cite{bao2024fastdetectgpt}
      & $.746_{\pm .000}$
      & $\mathbf{.046}_{\pm .001}$
      & $.124_{\pm .000}$ \\
    Log-Rank~\cite{su2023detectllm}
      & $.684_{\pm .000}$
      & $.060_{\pm .001}$
      & $.153_{\pm .000}$ \\
    MAGE~\cite{li-etal-2024-mage}
      & $.758_{\pm .000}$
      & $.159_{\pm .000}$
      & $.159_{\pm .000}$ \\
    RADAR~\cite{hu2023radar}
      & $.671_{\pm .000}$
      & $.191_{\pm .000}$
      & $.188_{\pm .000}$ \\
    MPU~\cite{tian2024multiscale}
      & $.637_{\pm .000}$
      & $.243_{\pm .000}$
      & $.248_{\pm .000}$ \\
    ReMoDetect~\cite{lee2024remodetect}
      & $.801_{\pm .000}$
      & $.050_{\pm .005}$
      & $.104_{\pm .000}$ \\
    RoBERTa-FT Classifier
      & $.733_{\pm .115}$
      & $.206_{\pm .069}$
      & $.188_{\pm .023}$ \\
    \midrule
    \textit{Ours}
      & $\mathbf{.809}_{\pm .025}$
      & $.120_{\pm .031}$
      & $\mathbf{.100}_{\pm .010}$ \\
    \bottomrule
  \end{tabular}
\end{table}

%% file: references.bib
@misc{comanici2025gemini25pushingfrontier,
      title={Gemini 2.5: Pushing the Frontier with Advanced Reasoning, Multimodality, Long Context, and Next Generation Agentic Capabilities}, 
      author={Gheorghe Comanici and others},
      year={2025},
      eprint={2507.06261},
      archivePrefix={arXiv},
      primaryClass={cs.CL},
      url={https://arxiv.org/abs/2507.06261}, 
}

@misc{openai2024gpt4technicalreport,
      title={GPT-4 Technical Report}, 
      author={OpenAI and others},
      year={2024},
      eprint={2303.08774},
      archivePrefix={arXiv},
      primaryClass={cs.CL},
      url={https://arxiv.org/abs/2303.08774}, 
}

@misc{grattafiori2024llama3herdmodels,
      title={The Llama 3 Herd of Models}, 
      author={Aaron Grattafiori and others},
      year={2024},
      eprint={2407.21783},
      archivePrefix={arXiv},
      primaryClass={cs.AI},
      url={https://arxiv.org/abs/2407.21783}, 
}

@inproceedings{bao2024fastdetectgpt,
  title     = {Fast-Detect{GPT}: Efficient Zero-Shot Detection of Machine-Generated Text via Conditional Probability Curvature},
  author    = {Guangsheng Bao and Yanbin Zhao and Zhiyang Teng and Linyi Yang and Yue Zhang},
  booktitle = {The Twelfth International Conference on Learning Representations},
  year      = {2024},
  url       = {https://openreview.net/forum?id=Bpcgcr8E8Z}
}

@misc{hans2024spotting,
  title         = {Spotting LLMs With Binoculars: Zero-Shot Detection of Machine-Generated Text},
  author        = {Abhimanyu Hans and Avi Schwarzschild and Valeriia Cherepanova and Hamid Kazemi and Aniruddha Saha and Micah Goldblum and Jonas Geiping and Tom Goldstein},
  year          = {2024},
  eprint        = {2401.12070},
  archiveprefix = {arXiv},
  primaryclass  = {cs.CL}
}

@inproceedings{li-etal-2024-mage,
  title     = {{MAGE}: Machine-generated Text Detection in the Wild},
  author    = {Li, Yafu  and
               Li, Qintong  and
               Cui, Leyang  and
               Bi, Wei  and
               Wang, Zhilin  and
               Wang, Longyue  and
               Yang, Linyi  and
               Shi, Shuming  and
               Zhang, Yue},
  editor    = {Ku, Lun-Wei  and
               Martins, Andre  and
               Srikumar, Vivek},
  booktitle = {Proceedings of the 62nd Annual Meeting of the Association for Computational Linguistics (Volume 1: Long Papers)},
  month     = aug,
  year      = {2024},
  address   = {Bangkok, Thailand},
  publisher = {Association for Computational Linguistics},
  url       = {https://aclanthology.org/2024.acl-long.3/},
  doi       = {10.18653/v1/2024.acl-long.3},
  pages     = {36--53}
}

@inproceedings{krishna2023paraphrasing,
  title     = {Paraphrasing evades detectors of {AI}-generated text, but retrieval is an effective defense},
  author    = {Kalpesh Krishna and Yixiao Song and Marzena Karpinska and John Frederick Wieting and Mohit Iyyer},
  booktitle = {Thirty-seventh Conference on Neural Information Processing Systems},
  year      = {2023},
  url       = {https://openreview.net/forum?id=WbFhFvjjKj}
}

@inproceedings{su2023detectllm,
  title     = {Detect{LLM}: Leveraging Log Rank Information for Zero-Shot Detection of Machine-Generated Text},
  author    = {Jinyan Su and Terry Yue Zhuo and Di Wang and Preslav Nakov},
  booktitle = {The 2023 Conference on Empirical Methods in Natural Language Processing},
  year      = {2023},
  url       = {https://openreview.net/forum?id=Dy2mbQIdMz}
}

@inproceedings{gehrmann-etal-2019-gltr,
  title     = {{GLTR}: Statistical Detection and Visualization of Generated Text},
  author    = {Gehrmann, Sebastian  and
               Strobelt, Hendrik  and
               Rush, Alexander},
  editor    = {Costa-juss{\`a}, Marta R.  and
               Alfonseca, Enrique},
  booktitle = {Proceedings of the 57th Annual Meeting of the Association for Computational Linguistics: System Demonstrations},
  month     = jul,
  year      = {2019},
  address   = {Florence, Italy},
  publisher = {Association for Computational Linguistics},
  url       = {https://aclanthology.org/P19-3019/},
  doi       = {10.18653/v1/P19-3019},
  pages     = {111--116}
}

@misc{patel2024lowresourceauthorshipstyletransfer,
      title={Low-Resource Authorship Style Transfer: Can Non-Famous Authors Be Imitated?}, 
      author={Ajay Patel and Nicholas Andrews and Chris Callison-Burch},
      year={2024},
      eprint={2212.08986},
      archivePrefix={arXiv},
      primaryClass={cs.CL},
      url={https://arxiv.org/abs/2212.08986}, 
}

@inproceedings{
wang2025humanizing,
title={Humanizing the Machine: Proxy Attacks to Mislead {LLM} Detectors},
author={Tianchun Wang and Yuanzhou Chen and Zichuan Liu and Zhanwen Chen and Haifeng Chen and Xiang Zhang and Wei Cheng},
booktitle={The Thirteenth International Conference on Learning Representations},
year={2025},
url={https://openreview.net/forum?id=PIpGN5Ko3v}
}

@misc{soto2025languagemodelsoptimizedfool,
      title={Language Models Optimized to Fool Detectors Still Have a Distinct Style (And How to Change It)}, 
      author={Rafael Rivera Soto and Barry Chen and Nicholas Andrews},
      year={2025},
      eprint={2505.14608},
      archivePrefix={arXiv},
      primaryClass={cs.CL},
      url={https://arxiv.org/abs/2505.14608}, 
}

@inproceedings{
nicks2024language,
title={Language Model Detectors Are Easily Optimized Against},
author={Charlotte Nicks and Eric Mitchell and Rafael Rafailov and Archit Sharma and Christopher D Manning and Chelsea Finn and Stefano Ermon},
booktitle={The Twelfth International Conference on Learning Representations},
year={2024},
url={https://openreview.net/forum?id=4eJDMjYZZG}
}

@inproceedings{
soto2024fewshot,
title={Few-Shot Detection of Machine-Generated Text using Style Representations},
author={Rafael Alberto Rivera Soto and Kailin Koch and Aleem Khan and Barry Y. Chen and Marcus Bishop and Nicholas Andrews},
booktitle={The Twelfth International Conference on Learning Representations},
year={2024},
url={https://openreview.net/forum?id=cWiEN1plhJ}
}

@inproceedings{ippolito-etal-2020-automatic,
    title = "Automatic Detection of Generated Text is Easiest when Humans are Fooled",
    author = "Ippolito, Daphne  and
      Duckworth, Daniel  and
      Callison-Burch, Chris  and
      Eck, Douglas",
    editor = "Jurafsky, Dan  and
      Chai, Joyce  and
      Schluter, Natalie  and
      Tetreault, Joel",
    booktitle = "Proceedings of the 58th Annual Meeting of the Association for Computational Linguistics",
    month = jul,
    year = "2020",
    address = "Online",
    publisher = "Association for Computational Linguistics",
    url = "https://aclanthology.org/2020.acl-main.164/",
    doi = "10.18653/v1/2020.acl-main.164",
    pages = "1808--1822"
}

@inproceedings{
lee2024remodetect,
title={ReMoDetect: Reward Models Recognize Aligned {LLM}'s Generations},
author={Hyunseok Lee and Jihoon Tack and Jinwoo Shin},
booktitle={The Thirty-eighth Annual Conference on Neural Information Processing Systems},
year={2024},
url={https://openreview.net/forum?id=pW9Jwim918}
}

@inproceedings{
tian2024multiscale,
title={Multiscale Positive-Unlabeled Detection of {AI}-Generated Texts},
author={Yuchuan Tian and Hanting Chen and Xutao Wang and Zheyuan Bai and QINGHUA ZHANG and Ruifeng Li and Chao Xu and Yunhe Wang},
booktitle={The Twelfth International Conference on Learning Representations},
year={2024},
url={https://openreview.net/forum?id=5Lp6qU9hzV}
}

@inproceedings{
hu2023radar,
title={{RADAR}: Robust {AI}-Text Detection via Adversarial Learning},
author={Xiaomeng Hu and Pin-Yu Chen and Tsung-Yi Ho},
booktitle={Thirty-seventh Conference on Neural Information Processing Systems},
year={2023},
url={https://openreview.net/forum?id=QGrkbaan79}
}

@misc{emi2024technicalreportpangramaigenerated,
      title={Technical Report on the Pangram AI-Generated Text Classifier}, 
      author={Bradley Emi and Max Spero},
      year={2024},
      eprint={2402.14873},
      archivePrefix={arXiv},
      primaryClass={cs.CL},
      url={https://arxiv.org/abs/2402.14873}, 
}

@inproceedings{
thai2026editlens,
title={EditLens: Quantifying the Extent of {AI} Editing in Text},
author={Katherine Thai and Bradley Emi and Elyas Masrour and Mohit Iyyer},
booktitle={The Fourteenth International Conference on Learning Representations},
year={2026},
url={https://openreview.net/forum?id=gOkitaPCfZ}
}

@inproceedings{
wu2024detectrl,
title={Detect{RL}: Benchmarking {LLM}-Generated Text Detection in Real-World Scenarios},
author={Junchao Wu and Runzhe Zhan and Derek F. Wong and Shu Yang and Xinyi Yang and Yulin Yuan and Lidia S. Chao},
booktitle={The Thirty-eight Conference on Neural Information Processing Systems Datasets and Benchmarks Track},
year={2024},
url={https://openreview.net/forum?id=ZGMkOikEyv}
}

@inproceedings{dugan-etal-2024-raid,
    title = "{RAID}: A Shared Benchmark for Robust Evaluation of Machine-Generated Text Detectors",
    author = "Dugan, Liam  and
      Hwang, Alyssa  and
      Trhl{\'i}k, Filip  and
      Zhu, Andrew  and
      Ludan, Josh Magnus  and
      Xu, Hainiu  and
      Ippolito, Daphne  and
      Callison-Burch, Chris",
    editor = "Ku, Lun-Wei  and
      Martins, Andre  and
      Srikumar, Vivek",
    booktitle = "Proceedings of the 62nd Annual Meeting of the Association for Computational Linguistics (Volume 1: Long Papers)",
    month = aug,
    year = "2024",
    address = "Bangkok, Thailand",
    publisher = "Association for Computational Linguistics",
    url = "https://aclanthology.org/2024.acl-long.674/",
    doi = "10.18653/v1/2024.acl-long.674",
    pages = "12463--12492"
}

@inproceedings{10.1007/978-3-032-04354-2_21,
author = {Bevendorff, Janek and Dementieva, Daryna and Fr\"{o}be, Maik and Gipp, Bela and Greiner-Petter, Andr\'{e} and Karlgren, Jussi and Mayerl, Maximilian and Nakov, Preslav and Panchenko, Alexander and Potthast, Martin and Shelmanov, Artem and Stamatatos, Efstathios and Stein, Benno and Wang, Yuxia and Wiegmann, Matti and Zangerle, Eva},
title = {Overview of~PAN 2025: Voight-Kampff Generative AI Detection, Multilingual Text Detoxification, Multi-author Writing Style Analysis, and~Generative Plagiarism Detection},
year = {2025},
isbn = {978-3-032-04353-5},
publisher = {Springer-Verlag},
address = {Berlin, Heidelberg},
url = {https://doi.org/10.1007/978-3-032-04354-2_21},
doi = {10.1007/978-3-032-04354-2_21},
booktitle = {Experimental IR Meets Multilinguality, Multimodality, and Interaction: 16th International Conference of the CLEF Association, CLEF 2025, Madrid, Spain, September 9–12, 2025, Proceedings},
pages = {388–411},
numpages = {24},
location = {Madrid, Spain}
}

@inproceedings{Platt1999ProbabilisticOF,
  title={Probabilistic Outputs for Support vector Machines and Comparisons to Regularized Likelihood Methods},
  author={John Platt},
  year={1999},
  url={https://api.semanticscholar.org/CorpusID:56563878}
}

@misc{glazer2025fewshotspeechdeepfakedetection,
      title={Few-Shot Speech Deepfake Detection Adaptation with Gaussian Processes}, 
      author={Neta Glazer and David Chernin and Idan Achituve and Sharon Gannot and Ethan Fetaya},
      year={2025},
      eprint={2505.23619},
      archivePrefix={arXiv},
      primaryClass={cs.SD},
      url={https://arxiv.org/abs/2505.23619}, 
}

@inproceedings{rivera-soto-etal-2021-learning,
    title = "Learning Universal Authorship Representations",
    author = "Rivera-Soto, Rafael A.  and
      Miano, Olivia Elizabeth  and
      Ordonez, Juanita  and
      Chen, Barry Y.  and
      Khan, Aleem  and
      Bishop, Marcus  and
      Andrews, Nicholas",
    editor = "Moens, Marie-Francine  and
      Huang, Xuanjing  and
      Specia, Lucia  and
      Yih, Scott Wen-tau",
    booktitle = "Proceedings of the 2021 Conference on Empirical Methods in Natural Language Processing",
    month = nov,
    year = "2021",
    address = "Online and Punta Cana, Dominican Republic",
    publisher = "Association for Computational Linguistics",
    url = "https://aclanthology.org/2021.emnlp-main.70/",
    doi = "10.18653/v1/2021.emnlp-main.70",
    pages = "913--919"
}

@inproceedings{10.5555/3295222.3295241,
author = {Geifman, Yonatan and El-Yaniv, Ran},
title = {Selective classification for deep neural networks},
year = {2017},
isbn = {9781510860964},
publisher = {Curran Associates Inc.},
address = {Red Hook, NY, USA},
booktitle = {Proceedings of the 31st International Conference on Neural Information Processing Systems},
pages = {4885–4894},
numpages = {10},
location = {Long Beach, California, USA},
series = {NIPS'17}
}

@misc{naeini2014binaryclassifiercalibrationnonparametric,
      title={Binary Classifier Calibration: Non-parametric approach}, 
      author={Mahdi Pakdaman Naeini and Gregory F. Cooper and Milos Hauskrecht},
      year={2014},
      eprint={1401.3390},
      archivePrefix={arXiv},
      primaryClass={stat.ML},
      url={https://arxiv.org/abs/1401.3390}, 
}

@inproceedings{wang-etal-2024-m4,
    title = "M4: Multi-generator, Multi-domain, and Multi-lingual Black-Box Machine-Generated Text Detection",
    author = "Wang, Yuxia  and
      Mansurov, Jonibek  and
      Ivanov, Petar  and
      Su, Jinyan  and
      Shelmanov, Artem  and
      Tsvigun, Akim  and
      Whitehouse, Chenxi  and
      Mohammed Afzal, Osama  and
      Mahmoud, Tarek  and
      Sasaki, Toru  and
      Arnold, Thomas  and
      Aji, Alham Fikri  and
      Habash, Nizar  and
      Gurevych, Iryna  and
      Nakov, Preslav",
    editor = "Graham, Yvette  and
      Purver, Matthew",
    booktitle = "Proceedings of the 18th Conference of the European Chapter of the Association for Computational Linguistics (Volume 1: Long Papers)",
    month = mar,
    year = "2024",
    address = "St. Julian{'}s, Malta",
    publisher = "Association for Computational Linguistics",
    url = "https://aclanthology.org/2024.eacl-long.83/",
    doi = "10.18653/v1/2024.eacl-long.83",
    pages = "1369--1407"
}

@misc{almazrouei2023falconseriesopenlanguage,
      title={The Falcon Series of Open Language Models}, 
      author={Ebtesam Almazrouei and Hamza Alobeidli and Abdulaziz Alshamsi and Alessandro Cappelli and Ruxandra Cojocaru and Mérouane Debbah and Étienne Goffinet and Daniel Hesslow and Julien Launay and Quentin Malartic and Daniele Mazzotta and Badreddine Noune and Baptiste Pannier and Guilherme Penedo},
      year={2023},
      eprint={2311.16867},
      archivePrefix={arXiv},
      primaryClass={cs.CL},
      url={https://arxiv.org/abs/2311.16867}, 
}

@InProceedings{pmlr-v38-hensman15,
  title = 	 {{Scalable Variational Gaussian Process Classification}},
  author = 	 {Hensman, James and Matthews, Alexander and Ghahramani, Zoubin},
  booktitle = 	 {Proceedings of the Eighteenth International Conference on Artificial Intelligence and Statistics},
  pages = 	 {351--360},
  year = 	 {2015},
  editor = 	 {Lebanon, Guy and Vishwanathan, S. V. N.},
  volume = 	 {38},
  series = 	 {Proceedings of Machine Learning Research},
  address = 	 {San Diego, California, USA},
  month = 	 {09--12 May},
  publisher =    {PMLR},
  url = 	 {https://proceedings.mlr.press/v38/hensman15.html}
}

@misc{mitchell2023detectgptzeroshotmachinegeneratedtext,
      title={DetectGPT: Zero-Shot Machine-Generated Text Detection using Probability Curvature}, 
      author={Eric Mitchell and Yoonho Lee and Alexander Khazatsky and Christopher D. Manning and Chelsea Finn},
      year={2023},
      eprint={2301.11305},
      archivePrefix={arXiv},
      primaryClass={cs.CL},
      url={https://arxiv.org/abs/2301.11305}, 
}

@inproceedings{
yang2024dnagpt,
title={{DNA}-{GPT}: Divergent N-Gram Analysis for Training-Free Detection of {GPT}-Generated Text},
author={Xianjun Yang and Wei Cheng and Yue Wu and Linda Ruth Petzold and William Yang Wang and Haifeng Chen},
booktitle={The Twelfth International Conference on Learning Representations},
year={2024},
url={https://openreview.net/forum?id=Xlayxj2fWp}
}

@inproceedings{verma-etal-2024-ghostbuster,
    title = "Ghostbuster: Detecting Text Ghostwritten by Large Language Models",
    author = "Verma, Vivek  and
      Fleisig, Eve  and
      Tomlin, Nicholas  and
      Klein, Dan",
    editor = "Duh, Kevin  and
      Gomez, Helena  and
      Bethard, Steven",
    booktitle = "Proceedings of the 2024 Conference of the North American Chapter of the Association for Computational Linguistics: Human Language Technologies (Volume 1: Long Papers)",
    month = jun,
    year = "2024",
    address = "Mexico City, Mexico",
    publisher = "Association for Computational Linguistics",
    url = "https://aclanthology.org/2024.naacl-long.95/",
    doi = "10.18653/v1/2024.naacl-long.95",
    pages = "1702--1717"
}

@inproceedings{artemova-etal-2025-beemo,
    title = "Beemo: Benchmark of Expert-edited Machine-generated Outputs",
    author = "Artemova, Ekaterina  and
      Lucas, Jason S  and
      Venkatraman, Saranya  and
      Lee, Jooyoung  and
      Tilga, Sergei  and
      Uchendu, Adaku  and
      Mikhailov, Vladislav",
    editor = "Chiruzzo, Luis  and
      Ritter, Alan  and
      Wang, Lu",
    booktitle = "Proceedings of the 2025 Conference of the Nations of the Americas Chapter of the Association for Computational Linguistics: Human Language Technologies (Volume 1: Long Papers)",
    month = apr,
    year = "2025",
    address = "Albuquerque, New Mexico",
    publisher = "Association for Computational Linguistics",
    url = "https://aclanthology.org/2025.naacl-long.357/",
    doi = "10.18653/v1/2025.naacl-long.357",
    pages = "6992--7018",
    ISBN = "979-8-89176-189-6"
}

@article{solaiman2019release,
  title={Release strategies and the social impacts of language models},
  author={Solaiman, Irene and Brundage, Miles and Clark, Jack and Askell, Amanda and Herbert-Voss, Ariel and Wu, Jeff and Radford, Alec and Krueger, Gretchen and Kim, Jong Wook and Kreps, Sarah and others},
  journal={arXiv preprint arXiv:1908.09203},
  year={2019}
}

@misc{liu2019robertarobustlyoptimizedbert,
      title={RoBERTa: A Robustly Optimized BERT Pretraining Approach}, 
      author={Yinhan Liu and Myle Ott and Naman Goyal and Jingfei Du and Mandar Joshi and Danqi Chen and Omer Levy and Mike Lewis and Luke Zettlemoyer and Veselin Stoyanov},
      year={2019},
      eprint={1907.11692},
      archivePrefix={arXiv},
      primaryClass={cs.CL},
      url={https://arxiv.org/abs/1907.11692}, 
}

@misc{gehring2025assessingllmtextdetection,
      title={Assessing LLM Text Detection in Educational Contexts: Does Human Contribution Affect Detection?}, 
      author={Lukas Gehring and Benjamin Paaßen},
      year={2025},
      eprint={2508.08096},
      archivePrefix={arXiv},
      primaryClass={cs.CL},
      url={https://arxiv.org/abs/2508.08096}, 
}

@InProceedings{pmlr-v5-titsias09a,
  title = 	 {Variational Learning of Inducing Variables in Sparse Gaussian Processes},
  author = 	 {Titsias, Michalis},
  booktitle = 	 {Proceedings of the Twelfth International Conference on Artificial Intelligence and Statistics},
  pages = 	 {567--574},
  year = 	 {2009},
  editor = 	 {van Dyk, David and Welling, Max},
  volume = 	 {5},
  series = 	 {Proceedings of Machine Learning Research},
  address = 	 {Hilton Clearwater Beach Resort, Clearwater Beach, Florida USA},
  month = 	 {16--18 Apr},
  publisher =    {PMLR},
  url = 	 {https://proceedings.mlr.press/v5/titsias09a.html}
}

@article{
sadasivan2025can,
title={Can {AI}-Generated Text be Reliably Detected? Stress Testing {AI} Text Detectors Under Various Attacks},
author={Vinu Sankar Sadasivan and Aounon Kumar and Sriram Balasubramanian and Wenxiao Wang and Soheil Feizi},
journal={Transactions on Machine Learning Research},
issn={2835-8856},
year={2025},
url={https://openreview.net/forum?id=OOgsAZdFOt},
note={}
}

@InProceedings{pmlr-v202-kirchenbauer23a,
  title = 	 {A Watermark for Large Language Models},
  author =       {Kirchenbauer, John and Geiping, Jonas and Wen, Yuxin and Katz, Jonathan and Miers, Ian and Goldstein, Tom},
  booktitle = 	 {Proceedings of the 40th International Conference on Machine Learning},
  pages = 	 {17061--17084},
  year = 	 {2023},
  editor = 	 {Krause, Andreas and Brunskill, Emma and Cho, Kyunghyun and Engelhardt, Barbara and Sabato, Sivan and Scarlett, Jonathan},
  volume = 	 {202},
  series = 	 {Proceedings of Machine Learning Research},
  month = 	 {23--29 Jul},
  publisher =    {PMLR},
  url = 	 {https://proceedings.mlr.press/v202/kirchenbauer23a.html}
}

@misc{he2024mgtbenchbenchmarkingmachinegeneratedtext,
      title={MGTBench: Benchmarking Machine-Generated Text Detection}, 
      author={Xinlei He and Xinyue Shen and Zeyuan Chen and Michael Backes and Yang Zhang},
      year={2024},
      eprint={2303.14822},
      archivePrefix={arXiv},
      primaryClass={cs.CR},
      url={https://arxiv.org/abs/2303.14822}, 
}
